\documentclass{article} 
\usepackage{iclr2025_conference,times}

\usepackage{amsmath,amsfonts,bm}

\def\eqref#1{equation~\ref{#1}}

\def\1{\bm{1}}

\DeclareMathAlphabet{\mathsfit}{\encodingdefault}{\sfdefault}{m}{sl}
\SetMathAlphabet{\mathsfit}{bold}{\encodingdefault}{\sfdefault}{bx}{n}

\usepackage{hyperref}
\usepackage{url}
\usepackage{multirow}
\usepackage{bigfoot}
\usepackage{adjustbox}
\usepackage{booktabs}
\usepackage{graphicx}   
\usepackage{subcaption} 
\usepackage{caption} 
\usepackage[colorinlistoftodos, textwidth=3cm]{todonotes}
\usepackage{comment}
\usepackage{float}

\title{Granite Embedding R2 Models}

\author{Granite Team \thanks{See Section \ref{sec:contributions} for full author list. For questions, comments, compliments contact awasthyp@us.ibm.com or raduf@us.ibm.com.} \thanks{For feedback or comments on this work, please open an issue at \url{https://github.com/ibm-granite/granite-embedding-models}.}\\
IBM Research AI
}

\iclrfinalcopy 
\begin{document}

\maketitle

\begin{abstract}

We introduce the Granite Embedding R2 models, a comprehensive family of high-performance English encoder-based embedding models engineered for enterprise-scale dense retrieval applications. Building upon our first-generation release, these models deliver substantial improvements, including 16x expanded context length (8,192 tokens), state-of-the-art performance across diverse retrieval domains - text, code, long-document search, multi-turn conversational, and tabular data - and measurable speed advantages of 19-44\% over leading competitors while maintaining superior accuracy. Our release encompasses both bi-encoder and cross-encoder architectures, featuring a highly effective 22-layer retriever model and its efficient 12-layer counterpart, alongside a high-quality reranker model, all trained exclusively on enterprise-appropriate data with comprehensive governance oversight. The models demonstrate exceptional versatility across standard benchmarks, IBM-developed evaluation suites, and real-world enterprise use cases, establishing new performance standards for open-source embedding models. In an era where retrieval speed and accuracy are paramount for competitive advantage, the Granite R2 models deliver a compelling combination of cutting-edge performance, enterprise-ready licensing, and transparent data provenance that organizations require for mission-critical deployments. All models are publicly available under the Apache 2.0 license at \url{https://huggingface.co/collections/ibm-granite}, enabling unrestricted research and commercial use.
\end{abstract}

\begin{figure}[h]
    \centering
    \includegraphics[width=.95\linewidth]{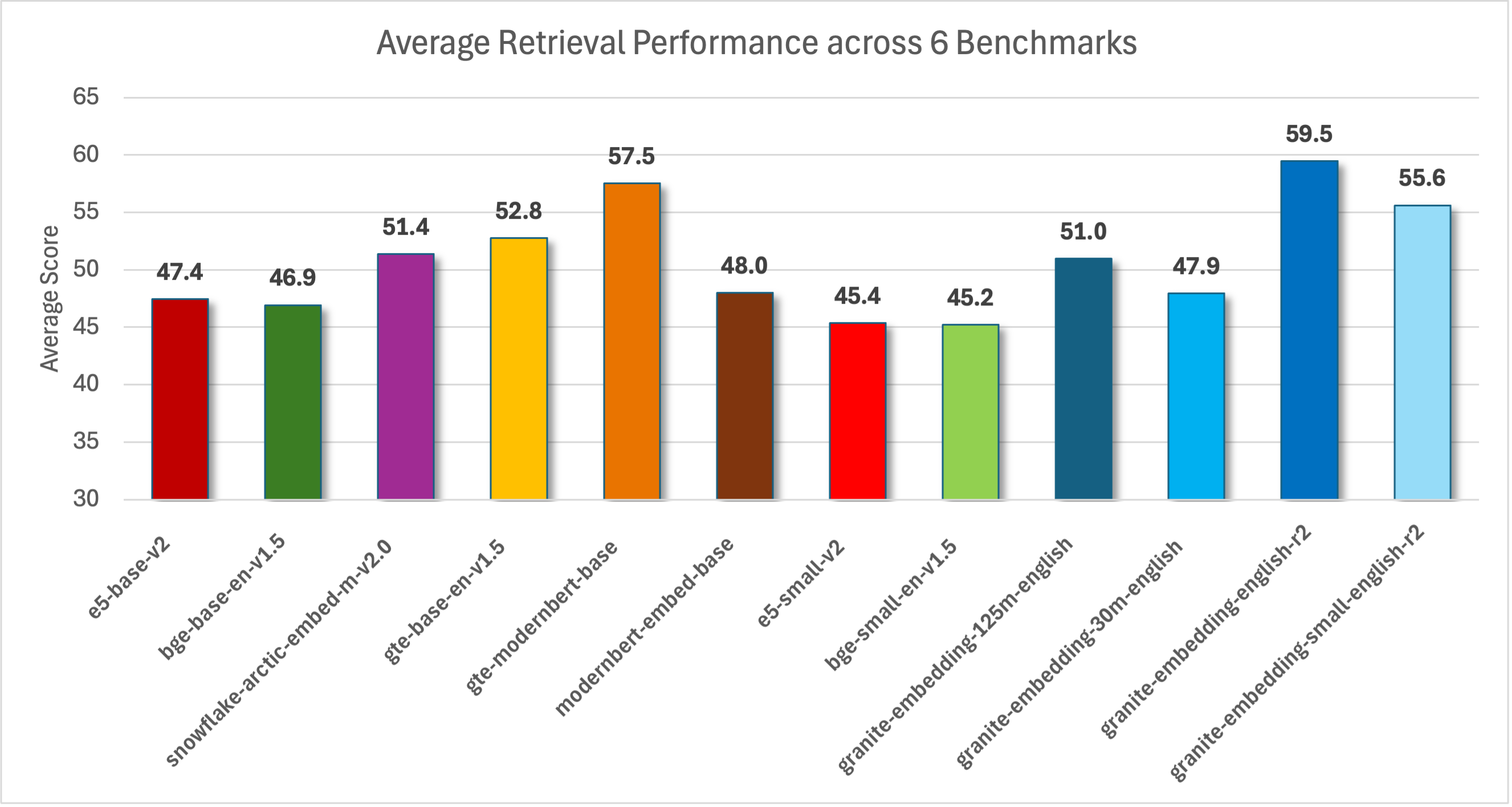}
    \caption{Performance of Granite R2 embedding models and comparable size open-source models, average over 6 retrieval benchmarks. Refer section \ref{subsec:retriever_performance} for details. }
    \label{fig:avg-perf}
\end{figure}

\section{Introduction}

Bi-encoder text embedding models convert text into a fixed-dimension vector, such that semantically close texts are close in the vector space, while dissimilar texts have a low similarity.  These embeddings can then be used in a variety of tasks, most commonly in retrieval applications, where the relevance of a document to a given query can be determined by the similarity of their embeddings \citep{dunn2017searchqa, xiong2020approximatenearestneighbornegative, neelakantan2022textcodeembeddingscontrastive, 10.1145/3269206.3271800, zhao2020spartaefficientopendomainquestion}, but also in document clustering \citep{angelov20} and text classification \citep{sun2019fine}. 

Encoder-based embedding models \citep{wang2022e5, bge_embedding, chen2024bgem3, merrick2024arcticembedv1, zhang2024mgte, nussbaum2024nomic} are widely used for these retrieval tasks, due to their low latency of inference and small memory footprint compared to decoder-based embedding models \citep{lee2024nvembed, wang2023e5mistral}. However, many encoder embedding models are often trained on data with non-commercial licenses, leading to restrictions in commercial deployment, and have a smaller context length, limiting their usability for long-context applications. 

While bi-encoders independently encode each text from a pair into fixed-length vectors that can be compared using distance measures like cosine similarity, cross-encoders or rerankers produce a single similarity score after jointly processing pairs of text. Cross-encoders often outperform bi-encoders, due to the ability of both texts to attend to each other, however, searching the entire corpus with them is computationally restrictive as it requires conducting an inference pass over every possible pair of query-document in the corpus \citep{reimers-gurevych-2019-sentence}. Thus, they are often used to rerank the top retrieved documents as judged by a bi-encoder, in a retrieve-and-rerank framework, that improves search quality without severe speed overhead.

This report introduces Granite Embedding R2 models, purpose built for information retrieval tasks, comprising of both bi-encoder and cross-encoder models. These models provide many improvements over our R1 Granite Embedding models \citep{awasthy2025granite}, including an increased context length, improved inference optimizations, and an updated encoder model based on the ModernBERT architecture \citep{warner2024modernbert}, trained on 2T tokens from a high-quality web-based corpus \citep{gohari2025gneissweb} and code data \citep{mishra2024granite}. The Granite Embedding models have been trained on high-quality, curated data, with data quality checks and screening to remove personal information and profane language. We release these models under the Apache 2.0 license, which is permissible for both commercial and research applications. We release three English models of varying sizes for a variety of inference budgets, spanning bi-encoder retriever models and cross-encoder rerankers:
\begin{itemize}
    \item \texttt{granite-embedding-english-r2} (149M parameters)\footnote{\href{https://huggingface.co/ibm-granite/granite-embedding-english-r2}{ibm-granite/granite-embedding-english-r2}}: with an output embedding size of 768, replacing granite-embedding-125m-english.
    \item \texttt{granite-embedding-small-english-r2} (47M parameters)\footnote{\href{https://huggingface.co/ibm-granite/granite-embedding-small-english-r2}{ibm-granite/granite-embedding-small-english-r2}}: a first-of-its-kind reduced-size model, with fewer layers and a smaller output embedding size (384), replacing granite-embedding-30m-english.
    \item \texttt{granite-embedding-reranker-english-r2} (149M parameters): a full-sized reranking model optimized for relevance ordering, built on granite-embedding-english-r2.
\end{itemize}

Both bi-encoders are designed to replace the older Granite embedding models, delivering state-of-the-art performance across standard and IBM-built information retrieval benchmarks (BEIR, ClapNQ), code retrieval (COIR), long-document search benchmarks (MLDR, LongEmbed), conversational multi-turn (MT-RAG), Table-IR (OpenWikiTables, NQTables, OTT-QA, MultiHierTT, AIT-QA), and on many enterprise use cases while supporting extended 8192-token context lengths meeting the industry standards. Crucially, the increase in parameter count for these models does not contribute to an increase in inference speed due to optimizations such as Flash Attention \citep{dao2023flashattention2}. 

Similarly, the accompanying reranker model demonstrates strong performance across diverse benchmarks, enabling sophisticated retrieve-and-rerank pipelines that maximize both recall and precision. Together, these models form a complete retrieval ecosystem that addresses the full spectrum of enterprise information retrieval challenges.

The paper's structure is as follows: Section \ref{sec:encoder} describes the training of the improved encoder models, which are based on the ModernBERT architecture. Section \ref{sec:biencoder} describes in detail the training recipes for the bi-encoder Granite retriever models, while Section \ref{sec:reranker} describes the training of the cross-encoder Granite Reranker model. Finally, Section \ref{sec:evaluation} presents a comprehensive evaluation of the Granite Embedding models, comparing their performance with that of other open-source encoder embedding models.

\section{Granite Encoder Models}
\label{sec:encoder}
The Granite Embedding R2 models have been trained on top of updated encoder models, featuring longer context lengths, a richer training corpus, and modern architectural improvements. We discuss the architecture of our base and small Granite Encoder models, which form the backbone of both the bi-encoder and cross-encoder models, as well as the training recipe based on \cite{warner2024modernbert} and details of the high-quality corpus used to optimize the model for code and long-context retrieval. 

\subsection{Encoder Model Architecture}

The Granite Encoder models have been trained following the ModernBERT \citep{warner2024modernbert} training recipe, including modern model optimizations such as alternating attention mechanism, rotary positional embeddings for flexible context length, and streamlined parameters. The models also support Flash Attention \citep{dao2023flashattention2} for improved efficiency, leading to no slowdown compared to the R1 models, even with a slightly larger parameter size, as shown in Section \ref{sec:time}. The Granite Encoder models use the ModernBERT tokenizer, which is a modified version of the OLMo tokenizer \citep{groeneveld2024olmo}, which shows better performance on code-related tasks. These encoder models have been trained on English and code data.

We train two models, \emph{granite-encoder-english} and \emph{granite-encoder-small-english}, which our our base and small sized models, the backbone of granite-embedding-english-r2 and granite-embedding-small-english-r2, respectively. The architecture of \emph{granite-encoder-english-r2} follows that of ModernBERT-base, with 22 layers, 149M parameters, and a vector size of 768. We select this architecture without any further ablations, referring to the findings of \cite{warner2024modernbert}. 
On the other hand, \emph{granite-encoder-small-english} is a first-of-its-kind small ModernBERT-\emph{style} model, with 12 layers, 47M parameters, and a vector size of 384, which still supports the increased context length of 8192 tokens. This architecture follows that of popular small embedding models \citep{li2023gte, bge_embedding, wang2022e5}, and was selected based on its performance on natural language understanding tasks \citep{wang-etal-2018-glue}, and downstream retrieval tasks \citep{kwiatkowski-etal-2019-natural-questions, husain2019codesearchnet, chen2024bgem3}. For ablations on the architectural choices of granite-encoder-small-english, please refer to Appendix \ref{app:small-encoder}. Similar to the ModernBERT architecture, our models have alternating global attention in every third layer. Detailed specifications of the architecture of each model are shown in Table~\ref{tab:architecture}.

\begin{table}[!t]
    \centering
    \setlength{\tabcolsep}{0.5em} 
    {\renewcommand{\arraystretch}{1.3}
    \begin{tabular}{l|cc}
    \hline
         & granite-encoder-small-english & granite-encoder-english\\ \hline
        Embedding size & $384$ & $768$ \\ 
        Layers & $12$ & $22$  \\  
        Intermediate size & $1536$ & $1152$  \\
        Global Rope Theta & $80000$ & $80000$  \\
        Vocabulary Size & $50368$  & $50368$ \\ 
        \hline
    \end{tabular}}
    \caption{Architectural Details for Granite Encoder Models}
    \label{tab:architecture}
\end{table}

\subsection{Training Data}
\label{sec:encoder-data}

We curate a diverse, high-quality corpus of text and code data to train our encoder models. The largest dataset we use in terms of tokens is the GneissWeb dataset \citep{gohari2025gneissweb}, a collection of web data filtered to create a high-quality corpora for language model training. We also include Wikipedia, BookCorpus, StackExchange, and PubMed articles in our training mix for diversity. For improved performance on Code-related tasks, we include a subset of Code data from the training corpora of Granite Code Models \citep{mishra2024granite}, also referred to here as Codepile. To improve performance on IBM benchmarks, we use internal IBM documents targeting specific technical domains. Furthermore, we also include multi-turn conversational data (Section~\ref{sec:retriever-data})r to improve performance on conversational IR tasks.

Through thorough ablations, we find that sampling from these datasets according to their relative sizes, with most of the weight being assigned to GneissWeb and CodePile,  yields better performance than a data curriculum in most cases. For the base model, we conduct the first two stages described in Section \ref{sec:encoder-training} on only data from GneissWeb, and conduct the last stage of training on a mixture of datasets. For the small model, we find mixing all datasets in all stages to be more beneficial, with the first two stages heavily sampling from GneissWeb.

\subsection{Training Recipe}
\label{sec:encoder-training}

Following the ModernBERT training setting, we train our models on the Masked Language Modeling objective over three distinct stages:

\begin{enumerate}
    \item Large Scale Pretraining: First, we train on 2 trillion tokens of text data, with a maximum context length of 1024. We use a Warmup-Stable-Decay learning rate schedule \citep{hu2024minicpm}, with peak learning rate of 8e-4 after an initial warmup on 3 billion tokens, and a RoPE theta of 10,000.
    \item Context Extension: We then scale up the context length to 8192 and the RoPE theta to 160,000 and train for 250 billion tokens on a constant learning rate of 3e-4.
    \item Learning Rate Decay: Finally, we train on 50 billion tokens, with the same context length and RoPE theta as above, but with a \texttt{1-sqrt} learning rate decay from the peak learning rate of 3e-4.
\end{enumerate}

We also use the StableAdamW optimizer \citep{wortsman2023stable}, and employ efficient training mechanisms such as sequence packing, unpadding, and flash attention, as described in \cite{warner2024modernbert}. We train both our base and small models from scratch, employing no special initialization techniques for either of the models.

\section{Granite Embedding R2}
\label{sec:biencoder}

Granite Embedding R2 models consist of high quality English embedding models, purpose built for retrieval tasks, using carefully curated enterprise-ready data. We discuss the training data and methodology for these models here, focusing on techniques such as retrieval oriented pretraining, contrastive finetuning and distillation.

\subsection{Training Data} 
\label{sec:retriever-data}

As detailed in Section~\ref{subsec: general training recipe}, the embedding models undergo a round of retrieval-oriented and tabular pretraining on general text data, and then contrastive finetuning on paired data. The data for each of these steps, including publically available and synthetically generated datasets, are discussed below.

\subparagraph{Retrieval Oriented Pretraining:} For RetroMAE-style pretraining \citep{xiao-etal-2022-retromae}, we train the model with about 200,000 sentences from Wikipedia, BookCorpus and StackExchange, with a context length of 8192 tokens. We experimented with other data mixtures, including those used for training the Granite Encoder models (Section~\ref{sec:encoder-data}), however, we find this data mix to give the best performance on downstream retrieval tasks.

\subparagraph{Tabular Pretraining:} Given the scarcity of paired retrieval training data and the need for tabular models to handle diverse structures and domains, we curate a corpus of approximately 8M tables from sources such as WikiTables \citep{openwikitables}, Arxiv Tables \citep{staghado_arxiv_tables2025}, PubTables \citep{pubtables}, Git Tables \citep{gittables}, FinTabNet \citep{fintabnet}, and NQ \citep{nqtables}. This collection provides a large and varied dataset for tabular pretraining, consisting of multiple formats, including CSV, Markdown, HTML, and table-marker representations to encourage format-agnostic learning.

A substantial portion of these tables contain numerical data, which poses challenges for common pretraining objectives such as RetroMAE \citep{xiao-etal-2022-retromae}, where masking often targets numbers, and the Inverse Cloze Task (ICT) \citep{ict-pretrain}, which relies on informative contextual text that is often unavailable or insufficient for tables. To address this, we generate synthetic summaries and metadata for the tables using Mistral-7B-Instruct \citep{mistral7b}, providing richer context for pretraining. In the table pretraining stage, we use both the original tables and the tables with summaries, and we also include a data replay from the RetroMAE pretraining stage.

\subparagraph{Retriever Training:}
Granite Embedding Models are trained on two types of paired data: 
\begin{enumerate}
    \item Weakly paired data mined from the web with in-batch negatives 
    \item High-quality annotated data with hard negatives for finetuning, from three key sources: 
    \begin{enumerate}
    \item Publicly available paired data
    \item IBM-internal paired data targeting specific technical domains 
    \item IBM-generated synthetic data.
    \end{enumerate}
\end{enumerate}
For governance, our data undergoes a data clearance process subject to technical, business, and governance review. This comprehensive process captures critical information about the data, including, but not limited to, its content description, intended use, data classification, licensing information, usage restrictions, as well as an assessment of sensitive information (i.e, personal information). 

The Granite Embedding R2 Models reuse training data from English Granite Embedding R1 Models \citep{awasthy2025granite}, and add data from the Code, Tabular, and Multi-Turn Conversation domains:
\begin{itemize}
    \item Code Data: We create code retrieval pairs from various sources to maintain diversity, mining hard negatives in most cases. We extract problem and solution pairs from Project CodeNet\footnote{\url{https://github.com/IBM/Project_CodeNet}}, consisting of pairs in Python, Java, C++ and C. The original problems, in mixed English and Japanese, were translated to English and further summarized to obtain a short query using Mixtral-8x22B. We also create pairs of code and natural language from the CoNaLa train set \citep{yin2018conala}, as well as code and docstring pairs extracted from CodePile \citep{mishra2024granite}. For CoNaLa and Codepile pairs, we mine hard negatives using granite-embedding-125m-english, while we select random negatives for Project CodeNet.
    \item Table-IR Data: We use several publicly available training datasets, including Open-WikiTables, NQTables, OTT-QA \citep{ottqa}, FinQA \citep{finqa}, and MultiHierTT \citep{multihiertt} for training \textit{granite-embedding-english-r2} for Table-IR tasks. For datasets that require query rewriting or de-contextualization, we implement a structured query-rewrite and filter pipeline using Mistral-7B-Instruct. To further enhance retrieval robustness, we mine hard negatives using granite-embedding-125m-english model and incorporate these during distillation from the teacher model, as described in Section \ref{subsec: general training recipe}.
    \item Multi-Turn Conversational IR Data:  For multi-turn conversationdal data, we use the train split of MultiDoc2Dal \citep{feng-etal-2021-multidoc2dial}. We also  synthetically generate about 2000 multi-turn conversations using Mixtral8x22B for the ClapNQ and IBM Cloud corpora of the MT-RAG dataset \citep{mtrag}. The generation involves grouping passages from the same document, generating turns (user query and assistant response) for each passage within the group and connecting these turns by de-contextualizing the queries.
\end{itemize}

\subsection{Training Recipe}
\label{subsec: general training recipe}

Embedding models are typically trained with a contrastive learning objective \citep{gao-etal-2021-simcse}, which brings the embeddings of a query closer to those of relevant passages and pushes them further away from non-relevant ones. Recent work \citep{zhang2024mgte, chen2024bgem3, li2023gte,bge_embedding,wang2022e5} employs a two-stage contrastive finetuning approach, first finetuning on a large corpus of semi-supervised pairs, then finetuning on a higher quality set of triples. The Granite Embedding models have been trained with additional techniques to improve performance, resulting in a training pipeline with the following stages:

\begin{enumerate}
    \item Retrieval Oriented Pre-training: starting with the Granite Encoder models, we conduct some steps of retrieval-oriented pre-training, such as RetroMAE \citep{xiao-etal-2022-retromae}, as a means to train the \texttt{[CLS]} vector to produce richer representation without explicitly training on the contrastive objective. This step is done with a context length of 8192 tokens.
    \item Tabular Pretraining: To effectively leverage large-scale tabular datasets for representation learning in retrieval applications, traditional bag-of-words techniques fall short due to cell redundancy and the prevalence of numerical content, which offers limited utility for learning correlations between tables and text \citep{tabert}. We therefore extend the RetroMAE framework to better learn tabular representations for retrieval. 
    
    Formally, let a table be represented as the token sequence $T = \bigl[ \texttt{[CLS]},\, h_1, \dots, h_J,\, \texttt{[SEP]},\, c_1, \dots, c_M,\, \texttt{[SEP]} \bigr],$ where \(h_i\) (\(i=1,\dots,J\)) are header tokens, and \(c_i\) (\(i=1,\dots,M\)) are cell tokens with \(M = m \times n\) for an \(m \times n\) table. Each table is paired with a natural language summary, represented as $S = \bigl[ \texttt{[CLS]},\, s_1, \dots, s_K \bigr],$ where \(s_i\) (\(i=1,\dots,K\)) are summary tokens. We apply an \mbox{M1 attention mask} \citep{m1attentionmask} on \(T\), and feed it into the encoder \(\mathbf{E}\) to generate a contextual table embedding via $z_T = \mathbf{E}\bigl(\widetilde{T}\bigr)_{\texttt{[CLS]}}.$ This representation $z_T$ is then provided to a shallow decoder $\mathbf{D}$, together with the input embeddings of the masked summary sequence $\widetilde{S}$. Unlike the original RetroMAE objective, where the decoder reconstructs the masked input sequence, our modified objective requires the decoder to predict masked tokens over the summary and table metadata, rather than the table tokens themselves:
    \[
        \mathcal{L}_{\text{Table-RetroMAE-Dec}}  = \sum_{i \in \mathcal{M}_S} \text{CE} (s_i\mid \mathbf{D}(z_T,  \widetilde{S}))
    \]
    where \(\mathcal{M}_S\) indexes the masked summary positions. We use a masking ratio of 20\% for the encoder and 60\% for the decoder tokens. This formulation forces the encoder to align table structure and content with textual summaries, enabling more effective representation learning for downstream table-text retrieval tasks.

    \item Contrastive Finetuning: the models are then finetuned for the contrastive learning objective on a large corpus of semi-supervised paired data, using the improved contrastive loss proposed in \citet{li2023gte}. Specifically, for a batch of triples $([q_i,(p_{ij})_j])_i$ consisting of a query and a set of passages -- without loss of generality, we can assume that $p_{i0}$ is a positive passage for query $i$, while $p_{ij}, j>0$ are negative passages -- we define the contrastive loss as:

    \[
        \mathcal{L}_{C} = -\frac{1}{n} \sum_{i=1}^{n} \mathrm{log} \frac{e^{s(q_i, p_{i0})}}{Z_i}
    \label{eq:cont-loss}
    \]
    \[
    \begin{aligned}
    Z_i = e^{s(q_i, p_{i0})} + \alpha\sum_{j>0} e^{s(q_i,p_{ij})} + \beta\sum_{i' \neq i} e^{s(q_i,q_{i'})} + \gamma\sum_{j>0} e^{s(p_{i0},p_{ij})}
    \label{eq:z-comp}
    \end{aligned}
    \]
    
    where $s(q, p)$ is a the temperature-scaled cosine similarity between the \texttt{[CLS]} embeddings of $q$ and $p$:
    \[
        s(q,p) = \frac{1}{\tau} \frac{\mathbf{E}(q)_\texttt{[CLS]} \cdot\mathbf{E}(p)_\texttt{[CLS]}} {\|\mathbf{E}(q)_\texttt{[CLS]}\|\|\mathbf{E}(p)_\texttt{[CLS]}\|}
    \label{eq:sim-score}
    \]

    Here, we can perform finetuning on a large set of semi-supervised pair data using a large batch size and in-batch negatives to better approximate the contrastive objective.
    
    \item Contrastive Distillation: Instead of further finetuning on high-quality triples, we instead distill the distribution of temperature-scaled cosine similarity scores from a Mistral-7B-Instruct model \citep{mistral7b} trained on the contrastive loss objective. Specifically, the training objective minimizes the cross entropy between the teacher's distribution $P_t$ of similarity scores between pairs and the student's distribution, $P_s$. Following \citet{hinton2014distilling}, we also scaled the score distribution of both teacher and student by a temperature, $\tau_{KD}$:
   \[
    \label{eq:kd-loss}
        \mathcal{L}_{KD} = - \sum_{i=1}^{n} \sum_{j=1}^{m} P_t(q_i, p_{ij}) \, \mathrm{log} \, P_s(q_i, p_{ij})
    \]
    \[
        P_s(q_i, p_{j}) =  \frac{e^{s_s(q_i, p_{ij})/\tau_{KD}}}{Z_{s,i}}
    \]
    \[
        P_t(q_i, p_{i0}) = \frac{e^{s_t(q_i, p_{ij})/\tau_{KD}}}{Z_{t,i}}
    \]

    We find this objective to yield a larger improvement in performance than finetuning with hard negatives. We use 3 mined hard negatives for more informative contrastive training, keeping the maximum sequence length of 1024 tokens.
    
    \item Domain Adaptation for Multi-turn Conversational IR: to improve the quality of the embeddings for multi-turn conversational IR, the models undergo a final domain-adaptation stage wherein the distribution of similarity scores from a domain-adapted Mistral teacher is distilled into the model. We conduct this step for granite-embedding-english-r2, and omit it for the small model as we do not see significant performance improvement in the latter case.
\end{enumerate}

At each stage of our training, we perform scaling of the global rope base frequency $\theta$ parameter for better Positional Interpolation \citep{meta-position-interpolation}. For both our models, we find using a global rope theta of 80K gives better performance than the default of 160K on both short and long-context downstream retrieval tasks, with ablations presented in Appendix \ref{app:rope-theta}.
Detailed hyperparameters for each stage are provided in Appendix~\ref{app:hyperparameters}. 

\subsection{Teacher Training}
\label{subsec:teacher-mistral}
We fine tune Mistral-7B-Instruct-v0.2 \citep{jiang2023mistral7b} using contrastive training as our teacher model for distillation. Similar to \citet{wang2024improvingtextembeddingslarge}, for each query-document pair $(q, p)$, we add an instruction template to the original query to generate a new one:
\[
    q_{inst} = Instruct: \texttt{\{task\_definition\}} Query: {q}
\]
where \texttt{\{task\_definition\}} is a placeholder for a one-sentence description of the embedding task. To calculate the embedding of a text, an \texttt{[EOS]} token is first appended to the end of the text before feeding it to the Mistral model, and the \texttt{[EOS]} vector of the last layer is treated as the text embedding.

We train two separate embedding models from the Mistral-7B-Instruct-v0.2 and create the final teacher model by merging the  two models together. The first model is trained using three million pairs of weakly paired data with in-batch negatives. The second model is trained using one million annotated data with one hard negative.

\section{Granite Reranker}
\label{sec:reranker}

The Granite reranker model is a cross-encoder model built on top of granite-embedding-english-r2, using a list-wise ranking objective. The model 
also features context length of 8192 tokens and is trained with curated high-quality data and carefully mined negatives.

\subsection{Training Recipe}
The granite-embedding-reranker-english-r2 is cross-encoder, that jointly encodes the query $q$ and document $d$ \ as $\bigl[ \texttt{[CLS]}\; q \; \texttt{[SEP]}\; d\;\bigr]$, and predicts the relevance score from the $\texttt{[CLS]}$ representation:
\[
    z = \text{Classifier}(h_{\texttt{[CLS]}})
\]

We randomly initialize the  classifier parameters, and fine-tune the model with PListMLE \citep{plistmle} loss objective, an extension of ListMLE \citep{listmle} which defines the probability of a permutation as the product of step-wise conditional probabilities. PListMLE biases the permutation probability distribution according to position-dependent weights, making it sensitive to rank positions. 

Given a ranked list $y = (y_1, y_2, \ldots, y_n)$ and relevance scores $z$, the loss is defined as:

\[
L(z, y) =
\sum_{i=1}^{n} \alpha(i) 
\left(
 -z_{y_i} 
 + \log \sum_{k=i}^{n} z_{y_k}
\right)
\]

where  $\alpha(\cdot)$ is a decreasing function. To align with NDCG, $\alpha(\cdot)$ is set as the gain function $2^{n-1}-1$, giving higher weight to more relevant documents.

We fine-tuned the model for 15K steps with a learning rate of 2e-4, weight decay 0.05, warmup ratio 0.15, and gradient-norm clipping at 1.0. We used the same training data as the Granite Embedding models, differing mainly in hard-negative mining and relative-order refinement. Hard negatives were first mined using \textit{granite-embedding-english-r2} by selecting negative documents whose query similarity is close to that of the positive documents (margin 0.95). From the retriever’s top-20 candidates, an in-house reranker, built on granite-embedding-125m-english, was then used to refine the ranking and produce a better-ordered negative set, ensuring that the selected negatives were harder and more semantically relevant. During training,  we used the top-8 hard negatives per query from this reranked list to optimize the ranking objective.
 
\section{Evaluation}
\label{sec:evaluation}

We evaluate the performance of our models on a variety of tasks and domain. Granite Encoder models are evaluated on a variety of natural language understanding and retrieval tasks in Appendix~\ref{app:encoder-performance}. 

We evaluate Granite Embedding models on retrieval benchmarks across domains, such as text, code, table, conversational, and long context retrieval in in \ref{subsec:retriever_performance} and \ref{sec:time}. The models show strong performance, achieving higher scores than other open source models on average, while maintaining the highest inference speeds 

We evaluate Granite Reranker model by reranking the top-20 documents retrieved by granite-embedding-english-r2 and granite-embedding-small-english-r2 on various retrieval datasets in \ref{subsec:reranker_performance} and show the reranker model shows strong performance over other open-source rerankers. 

\subsection{Retrieval Performance}
\label{subsec:retriever_performance}

We evaluate the Granite Embedding models on a variety of retrieval tasks, spanning multiple domains, document lengths and text objects (eg. documents, tables, conversations):
\begin{itemize}
    \item English Retrieval: We evaluate on general information retrieval benchmarks such as  \citep{enevoldsen2025mmtebmassivemultilingualtext}, comprising retrieval tasks on a variety of domains with a focus on zero-shot evaluations. We also include evaluation on the popular BEIR benchmark \citep{thakur2021beir} in Appendix~\ref{app:evals}.
    \item Code Retrieval: We evaluate on code retrieval tasks of the COIR benchmark \citep{li2024coircomprehensivebenchmarkcode}, which consists of text-to-code, code-to-text, and hybrid code retrieval.
    \item Long Context Retrieval: To evaluate performance on retrieving long-context documents, we measure the performance on the MLDR task \citep{chen2024bgem3} and the LongEmbed benchmark \citep{zhu2024longembed}.
    \item Table Retrieval: Existing text embedding models frequently underperform when encoding structured data such as tables \citep{strubert22, mouravieff-etal-2025-structural}. We evaluate on the tabular retrieval task across five datasets: OpenWikiTables, NQTables, OTT-QA, MultiHierTT, and AIT-QA \citep{aitqa}.
    \item Multi-Turn Conversation Retrieval: We evaluate our models to retrieve documents in a multi-turn conversation setting using the MT-RAG retrieval task \citep{katsis2025mtrag}.
\end{itemize}

We compare our bi-encoders with other state-of-the-art embedding models of similar parameter size. granite-embedding-enlish-r2 is compared to popular open source base models, such as BGE Base \citep{bge_embedding}, E5 Base \citep{wang2022e5}, as well as recent models with larger sequence length, such as Arctic Embed (M) \citep{yu2024arcticembedv2}, GTE Base \citep{zhang2024mgte,li2023gte}, GTE ModernBERT Base \citep{zhang2024mgte,li2023gte}, and Nomic-AI ModernBERT Embed Base \citep{nussbaum2024nomic}. The small model is compared to BGE Small \citep{bge_embedding} and E5 Small \citep{wang2022e5}, and is one of the first small models with a long context. We also compare the R2 embedding models to the R1 Granite Embedding Models \citep{awasthy2025granite}, to quantify the improvement over the previous release.

\begin{table}[!t]
    \centering
    \begin{adjustbox}{width=1.0\textwidth}
    {\renewcommand{\arraystretch}{1.3}
    \begin{tabular}{l|c|c|c|c|c|c|c|c|c}
    \toprule
        Model                               &  Params  &  Embed.  & Avg. & MTEB-v2 &  CoIR &  MLDR  & LongEmbed & Table IR & MTRAG \\ 
        & (M) & Size &  & Retrieval (10)  &  (10)  & (En)  & (6) & (5) & (4) \\
        \hline
        e5-base-v2                          & 109 & 768 & 47.4 & 49.7 & 50.3 & 32.5 & 41.1 & 74.1 & 37.0 \\ 
        bge-base-en-v1.5                    & 109 & 768 & 46.9 & 54.8 & 46.6 & 33.5 & 33.9 & 74.0 & 38.8 \\ 
        snowflake-arctic-embed-m-v2.0       & 305 & 768 & 51.4 & 58.4 & 52.2 & 32.4 & 55.4 & 80.8 & 29.2 \\ 
        gte-base-en-v1.5                    & 137 & 768 & 52.8 & 55.5 & 42.4 & 42.7 & 59.4 & 80.5 & 36.0 \\ 
        gte-modernbert-base                 & 149 & 768 & 57.5 & 57.0 & 71.5 & 46.2 & 57.0 & 76.7 & 36.8 \\ 
        nomic-ai/modernbert-embed-base      & 149 & 768 & 48.0 & 48.7 & 48.8 & 31.3 & 56.3 & 66.7 & 36.2 \\ 
        e5-small-v2                         & 33 & 384 & 45.4 & 48.5 & 47.1 & 29.9 & 40.7 & 72.3 & 33.8 \\ 
        bge-small-en-v1.5                   & 33 & 384 & 45.2 & 53.9 & 45.8 & 31.4 & 32.1 & 69.9 & 38.2 \\ 
        \midrule
        granite-embedding-125m-english      & 125 & 768 & 51.0 & 55.7 & 50.3 & 35.0 & 42.2 & 73.3 & 49.5 \\ 
        granite-embedding-30m-english       & 30 & 384 & 47.9 & 51.8 & 47.0 & 32.6 & 38.4 & 69.2 & 48.6 \\ 
        \textbf{granite-embedding-english-r2}  & 149 & 768 & \textbf{59.5} & 56.4 & 54.8 & 41.6 & 67.8 & 78.5 & 57.6 \\ 
        \textbf{granite-embedding-small-english-r2} & 47 & 384 & \textbf{55.6} & 53.9 & 53.4 & 40.1 & 61.9 & 75.5 & 48.9 \\ 
    \bottomrule
    \end{tabular}}
    \end{adjustbox}
    \caption{Retrieval Performance. Average scores are reported for benchmarks, with the number of tasks indicated in parentheses. MTRAG shows Recall@5. Two datasets in LongEmbed use Accuracy@1. Three tasks in Table IR use Recall@5 and two use Match@5. All other scores are average NDCG@10. Complete breakdown of scores is provided in Appendix \ref{app:evals}. }
\label{tab:performance}
\end{table}

Unless specifically mentioned, all tasks are evaluated with a maximum sequence length of 8192. While we show only the average performance for each benchmark in Table \ref{tab:performance}, we give a detailed evaluation for our models in Appendix \ref{app:evals}, including the evaluation on the complete English MTEB-v2 benchmark in Appendix \ref{app:mteb-v2}. 

As shown in Table \ref{tab:performance}, Granite embedding R2 models show a strong performance across diverse tasks despite all tasks being \textbf{zero-shot} except for NQ, Hotpot, FEVER. Notably, on average, granite-embedding-english-r2 outperforms other models of similar size.  The Granite Embedding R2 models achieve state-of-the-art performance on long-context retrieval benchmarks like LongEmbed, with granite-embedding-english-small-r2 having a very high accuracy (even compared to some larger models) without increasing inference cost. 

\subsection{Embedding Speed}
\label{sec:time}

Text embedding models are fundamental to information retrieval systems and Retrieval-Augmented Generation (RAG) applications. Organizations typically process millions of documents, with frequent updates and new content requiring continuous ingestion. This makes encoding speed as important as accuracy—a slow model can become a significant bottleneck in large-scale deployments.

\begin{table}[!ht]
    \centering
    {\renewcommand{\arraystretch}{1.3}
    \begin{tabular}{l|c|c|c|c}
    \toprule
         Data & Range  & Avg. & Range  & Avg. \\
         & (chars) & (chars) & (tokens) & (tokens) \\
         \midrule
         IBM documentation & [10, 475001] & 6393 & [2, 329116] & 1873 \\
    \bottomrule
    \end{tabular}}
    \caption{Data used for speed computations. Token size is determined with the BERT tokenizer (used by the e5-base-v2 model)}
    \label{tab:speed_data}
\end{table}

To evaluate the performance of embedding models in realistic scenarios, we construct a benchmark using 23,000 public IBM technical documents covering various products from ServeRAID controllers to IBM Storwize systems. Guiding principles for the experiment:
\begin{itemize}
    \item Create a realistic scale and document complexity: a large number of documents of varying lengths,
    ranging from 10 to 475,001 characters (averaging 6,393 characters - see Table \ref{tab:speed_data})
    \item Use standardized processing: 512-token chunks with 100-token overlaps across all models
    \item Perform consistent testing: Identical corpus and hardware (single Nvidia H100 GPU) for all tests
\end{itemize}

\begin{table}[!t]
    \centering
    \begin{adjustbox}{width=1\textwidth}
    {\renewcommand{\arraystretch}{1.3}
    \begin{tabular}{l|c|c|c|c}
    \toprule
        Model                               &  Parameters  &  Embedding Size  & Encoding Speed  & Rel to Granite \\ 
         &  (M) $\downarrow$ &    & (Docs/s) $\uparrow$  &  R2 equivalent\\ 
        \midrule
        e5-base-v2                          & 109 & 768 & 115 & -20.1\%\\ 
        bge-base-en-v1.5                    & 109 & 768 & 116 & -19.4\%\\ 
        snowflake-arctic-embed-m-v2.0       & 305 & 768 & 73  & -49.3\%\\ 
        gte-base-en-v1.5                    & 137 & 768 & 116 & -38.9\%\\ 
        gte-modernbert-base                 & 149 & 768 & 88  & -1.4\%\\ 
        nomic-ai/modernbert-embed-base      & 149 & 768 & 87  & -2.2\%\\ 
        granite-embedding-125m-english      & 125 & 768 & 149 & 3.5\% \\ 
        \textbf{granite-embedding-english-r2}  & 149 & 768 & 144 & 0\% \\ 
        \midrule
        e5-small-v2                         & 33 & 384 & 138  & -44.2\%\\ 
        bge-small-en-v1.5                   & 33 & 384 & 138  & -44.2\%\\ 
        granite-embedding-30m-english       & 30 & 384 & 198  & -0.5\% \\ 
        \textbf{granite-embedding-small-english-r2}  & 47 & 384 & 199 & 0\% \\ 
    \bottomrule
    \end{tabular}}
    \end{adjustbox}
    \caption{Retrieval time in terms of encoding speed and ingestion time. All evaluations are done on a single Nvidia H100 GPU, with a batch size of 128. The last column represents the relative speed of the given model to the size-equivalent granite R2 model (either base or small)}
\label{tab:time}
\end{table}

As seen in Table \ref{tab:time}, the Granite R2 embedding models perform well across several metrics:
\begin{itemize}
    \item Speed performance: fast with high performance 
    \begin{itemize}
        \item granite-embedding-small-english-r2 processes 199 documents per second, which is 38\% faster than comparable ModernBERT models (144 docs/s)
        \item granite-embedding-english-r2 processes 144 documents per second, outperforming smaller BERT-based alternatives like e5-small-v2 and bge-small-en-v1.5 (both at 138 docs/s)
        \item The larger granite models consistently outperform smaller competitors, indicating good architectural efficiency
    \end{itemize}

    \item Model efficiency: The Granite R2 series maintains the speed advantages of their R1 predecessors while offering additional capabilities such as an increased context length. Despite architectural changes, both models run with the same speed as their predecessors.

    \item Comparison with alternatives: When compared against popular open-source alternatives like bge-base-en-v1.5, granite models show competitive performance.
\end{itemize}

\subsection{Reranking Performance}
\label{subsec:reranker_performance}

We evaluate the Granite Reranker model on BEIR, MLDR and Miracl benchmarks, a common benchmark used for general information retrieval. We compare our reranker model with other state-of-the-art ranking models 
such as MiniLM-L12, BGE Base and Large \citep{bge_embedding}, and GTE base \cite{zhang2024mgte}. All models are evaluated on the top-20 documents retrieved by granite-embedding-english-r2. Each reranking model is evaluated with its maximum supported sequence length, while queries are truncated to 64 tokens.

\begin{table}[!ht]
    \centering
    \small
    {\renewcommand{\arraystretch}{1.3}
    \begin{tabular}{l|c|c|c|c|c}
    \toprule
        Model                                   &  Parameters   & Seq.      & BEIR   & MLDR  & Miracl \\ 
                                                & (M)           & Length    & Avg.   & (en)  & (en) \\\hline
        
        Retriever: granite-embedding-small-english-r2  &  47           &    8192   & 50.9   & 40.1  & 42.4 \\ \midrule
        \hspace*{5mm}ms-marco-MiniLM-L12-v2                  &  33           &    512    & 52.0   & 34.8  & 54.5  \\
        \hspace*{5mm}bge-reranker-base                       & 278           &    512    & 51.6   & 36.7  & 40.7 \\ 
        \hspace*{5mm}bge-reranker-large                      & 560           &    512    & 53.0   & 37.9  & 42.2\\ 
        \hspace*{5mm}gte-reranker-modernbert-base            & 149           &    8192   & 54.8   & 51.2  & 54.3 \\
        \hspace*{5mm}\textbf{granite-embedding-reranker-english-r2 }               & 149           &    8192   & 54.4   & 44.9  & 53.7 \\ 
    \midrule        
        Retriever: granite-embedding-english-r2& 149           &    8192   & 53.1   & 41.6  & 43.6  \\ \midrule
        \hspace*{5mm}ms-marco-MiniLM-L12-v2                  &  33           &    512    & 53.2   & 34.5  & 55.4  \\
        \hspace*{5mm}bge-reranker-base                       & 278           &    512    & 53.0   & 36.6  & 40.9  \\ 
        \hspace*{5mm}bge-reranker-large                      & 560           &    512    & 54.3   & 38.0  & 42.3  \\ 
        \hspace*{5mm}gte-reranker-modernbert-base            & 149           &    8192   & 56.1   & 50.4  & 54.8  \\ 
        \hspace*{5mm}\textbf{granite-embedding-reranker-english-r2}                & 149           &    8192   & 55.4   & 44.4  & 54.5  \\ 
    \bottomrule
    \end{tabular}}
    \caption{Reranking performance on BEIR, MLDR and Miracl benchmarks. All scores are average NDCG@10. All the reraranking models rerank top-20 outputs from the granite-embedding-english-small-r2 and granite-embedding-english-r2, retriever respectively}
\label{tab:rankingGranite}
\end{table}

As shown in Table \ref{tab:rankingGranite}, the Granite Reranker model achieve strong performance on all benchmarks, improving the performance over using the retriver alone, while outperforming all reranker models except \cite{zhang2024mgte}. The gap with \cite{zhang2024mgte} is relatively small, with the main difference observed on MLDR. Notably, \cite{zhang2024mgte} incorporates MS MARCO and MLDR in training, whereas we do not.

\section{Conclusion}
In this work, we have presented the Granite Embedding R2 Models, a family of specialized retrieval and reranker models designed to address the computational and accuracy requirements of enterprise-scale information retrieval systems. Our experimental evaluation demonstrates that these models achieve substantial performance gains, with processing speeds that are 19\% faster than leading base model baselines and 44\% faster than competitive small model alternatives, while preserving state-of-the-art retrieval accuracy across diverse domains including text, code, conversational data, tabular content, and long-context scenarios.

The proposed models incorporate several key contributions: optimized bi-encoder and cross-encoder architectures that support extended 8192 token contexts, comprehensive data curation methodologies ensuring enterprise-grade quality standards, and transparent training procedures. We release these models under the Apache 2.0 license supporting both academic research and practical deployment scenarios.
Our findings indicate that the Granite R2 model family provides a viable solution for organizations seeking to implement robust, scalable information retrieval systems. The combination of computational efficiency, retrieval performance, and enterprise-ready licensing positions these models as a significant contribution to both academic research and business-critical applications.

In an era where milliseconds matter and accuracy cannot be compromised, Granite R2 models don't just meet the standard—they set it.

\bibliography{main}
\bibliographystyle{iclr2025_conference}

\newpage
\appendix

\section{Contributions}
\label{sec:contributions}
The Granite R2 embedding models were truly the outcome of a successful collaboration across
geographies led by Radu Florian - with contributions from IBM Watson Research Lab (WRL) lab
and India Research Lab (IRL). Parul Awasthy was the challenge lead on the project overall, calling from WRL, with Jaydeep Sen
coordinating the work from IRL. We are very grateful for the wonderful and successful collaboration
across continents - looking forward to even better models!

\subsubsection*{Encoder Model Training}
Parul Awasthy, Aashka Trivedi

\subsubsection*{Retriever and Reranker Training}
Parul Awasthy, Riyaz Bhat, Meet Doshi, Bhavani Iyer, Vishwajeet Kumar, Yulong Li, Vignesh P, Aashka Trivedi, Yushu Yang

\subsubsection*{Data and Evaluation}
Parul Awasthy, Ken Barker, Meet Doshi, Radu Florian, Martin Franz, Bhavani Iyer, Vishwajeet Kumar, Yulong Li, Rudra Murthy, Vignesh P, Aashka Trivedi, Todd Ward

\subsubsection*{Product Management}
Abraham Daniels, Madison Lee

\subsubsection*{Technical Leadership}
Parul Awasthy, David Cox, Radu Florian, Luis Lastras, Salim Roukos, Jaydeep Sen

\section{granite-encoder-small-english Architecture Ablations}
\label{app:small-encoder}

\begin{table}[!ht]
    \centering
    \begin{adjustbox}{width=1\textwidth}
    {\renewcommand{\arraystretch}{1.3}
    \begin{tabular}{l|l|c|c|ccc}
    \toprule
        Description & Architecture & Params  & NLU & \multicolumn{3}{c}{Retrieval}  \\ 
        ~ & ~ & (M) $\downarrow$ & GLUE & CSN & NQ & MLDR \\ \midrule
        granite-encoder-30m-english & L6-H384-I1536-A12 & 30 & 79.8 & 38.1 & 28.7 & 13.4 \\ \midrule
        Half Layers & L12-H384-I1152-A12 & 42 & 80.4 & 66.1 & 34.3 & 23.9  \\ 
        Half Layers, FFN & L12-H384-I576-A12 & 34 & 78.9 & 65.3 & 32.9 & 20.9 \\ 
        Half Layers, Attention  & L12-H384-I1152-A6 & 42 & 80.6 & 65.6 & 33.6 & 11.3  \\ 
        Quarter Layers & L6-H384-I1152-A12 & 31 & 76.9 & 60.9 & 29.4 & 13.6  \\ 
        Half FFN & L22-H384-I576-A12 & 47 & 81.1 & 66.6 & 34.9 & 27.5  \\ \midrule
        Maintain base-large ratio  & L18-H576-I576-A9 & 71 & 83.4 & 69.7 & 37.6 & 23.1 \\ 
        Base/Large ratio; HS384; head size 64 & L18-H384-I576-A6 & 42 & 81.1 & 67.1 & 33.7 & 21.8 \\ \midrule
        Half FFN; layer = 3x+1 & L16-H384-I576-A12 & 39 & 80.2 & 66.6 & 33.8 & 23.7  \\ 
       \multirow{2}{5cm}{Base/Large ratio; HS384; head size 64; layer = 3x+1} & L19-H384-I576-A6 & 43 & 81.3 & 66.9 & 34.4 & 18.7  \\ 
         & L16-H384-I576-A6 & 39 & 80.3 & 64.4 & 33.8 & 11.6  \\ \midrule
        Standard small architecture & L12-H384-I1536-A12 & 47 & 80.5 & 68.3 & 35.0 & 25.6  \\ 
        \bottomrule
    \end{tabular}}
    \end{adjustbox}
    \caption{Architecture Ablations for granite-encoder-small-english. Architecture depicted as Layers-HiddenSize-IntermediateSize-AttentionHeads.  Models with a hidden size of 576 are included for completeness, however only models with hidden sizes of 384 are considered for selection.}
\label{tab:small-encoder-ablation}
\end{table}

granite-encoder-small-english is one of the first open small ModernBERT-style encoder model, and we explore the following options for its architecture, loosely based on the ModernBERT-base architecture:
\begin{enumerate}
    \item Modifications to ModernBERT-base architectures: we experiment with halving the number of layers, intermediate size and the attention size of the ModernBERT-base architecture, keeping other dimensions the same. We also try to take a fourth of the number of layers to further reduce latency of inference. 
    \item Maintain the ratio of dimensions of ModernBERT-base to ModernBERT-large: the ModernBERT models come in both base and large varieties. We create architectures of small models such that the ratio of small-to-base dimensions maintain that of base-to-large. Note, the model architecture that maintains this ratio in all dimensions has a hidden size of 576- we include the performance of this model for completeness, however we intend to choose an architecture which produces embeddings of size 384, as a replacement to granite-embedding-30m-english.
    \item Attention head size of 64: An important aspect of the ModernBERT architecture is that the attention head size (i.e., the hidden size divided by the number of attention heads) is 64. We experiment with maintaining this ratio.
    \item Maintaining the number of layers to be of the form $3x + 1$: the ModernBERT architecture alternates between global and local attention, keeping global attention at every third layer. We fix this design in both our base and small encoder models, and thus ablate the effects of maintaining a $3x+1$ pattern for the small model, such that the last layer always has global attention.
    \item Standard ``small" architecture: Finally, we use the standard architecture of many small embedding models, with 12 layers and a hidden size of 384.
\end{enumerate}

We ablate the architecture for granite-encoder-small-english by training various choices on 100B tokens for the first stage of training (Section \ref{sec:encoder-training}). As a baseline, we choose the encoder used to train granite-embedding-30m-english, a RoBERTa-like small model, which we refer to as granite-encoder-30m-english.  We then evaluate the performance of the encoder on natural language understanding tasks of the GLUE benchmark \cite{wang-etal-2018-glue}. For estimating downstream retrieval performance, we finetune each encoder for 1 epoch on MS-Marco triples with a contrastive learning objective, and evaluate the performance of the resulting embedding models on general retrieval (Natural Questions \cite{kwiatkowski-etal-2019-natural-questions}), code retrieval (CodeSearchNet \citep{husain2019codesearchnet}), and long-context retrieval (MLDR \citep{chen2024bgem3}). 

As shown in Table \ref{tab:small-encoder-ablation}, the standard small encoder gives the best performance on downstream retrieval tasks with a vector size of 384, without much loss in Glue Performance and a modest increase in parameter size compared to other low-performing ablations. We thus select this architecture as the basis of granite-enoder-english. 

\section{Encoder Performance}
\label{app:encoder-performance}

\begin{table}[!ht]
    \centering
    \begin{adjustbox}{width=1.0\textwidth}
    {\renewcommand{\arraystretch}{1.3}
    \begin{tabular}{l|lllllllll|lll}
    \toprule
        Model & \multicolumn{9}{c|}{GLUE} & \multicolumn{3}{c}{Retrieval}  \\ \hline
        ~ & MNLI & QNLI & QQP & SST2 & CoLA & RTE & MRPC & STSB & Avg. & CSN & NQ & MLDR \\ \hline
        BERT Base & 84.8 & 	91.4 & 	88.3 & 	93.2 & 	55.1 & 	75.2 & 	91.4 & 	90.0 & 	83.7 & 52.5 & 	44.5 & 	23.5 \\ 
        RoBERTa Base & 87.5 & 	92.8 & 	88.7 & 	94.0 & 	60.2 & 	84.8 & 	92.1 & 	90.9 & 	86.4 & 63.5	& 42.0 &	19.1 \\ 
        ModernBERT Base & 88.8 & 93.0 & 89.2 & 95.2 & 63.0 & 87.5 & 91.9 & 91.3 & 87.5 & 76.6 & 45.6 & 27.5 \\ 
        MiniLM-L12-H384 & 84.9 & 	91.1 & 	88.0 & 	92.0 & 	53.3 & 	76.5 & 	91.2 & 	90.5 & 	83.4 & 50.7 & 39.5  & 19.4 \\ \midrule
        granite-encoder-english & 88.4 & 92.8 & 89.0 & 94.4 & 62.2 & 84.2 & 92.0 & 90.9 & 86.8 & 69.4 & 45.7 & 31.5 \\ 
        granite-encoder-small-english & 83.7 & 89.0 & 87.3 & 92.4 & 50.8 & 71.7 & 90.3 & 87.2 & 81.6 & 69.1 & 36.1 & 25.5 \\ 
    \bottomrule
    \end{tabular}}
    \end{adjustbox}
    \caption{Encoder Performance on NLU and Retrieval Tasks. Retrieval performance measured as NDCG@10 scores after finetuning on MS-MARCO Triples \citep{bajaj2018msmarco} }
\label{tab:encoder-performance}
\end{table}

We evaluate our encoder models on natural language understanding and downstream retrieval performance:
\begin{itemize}
    \item NLU Performance: for natural language understanding, we measure the performance on the GLUE benchmark \cite{wang-etal-2018-glue}, comprising 8 language understanding tasks. Following \cite{warner2024modernbert}, for the RTE, MRPC, and STS-B tasks, we start training from the MNLI checkpoint.
    \item Retrieval Performance: The Granite Encoder models are purpose-built for retrieval tasks, and we evaluate the dense retrieval performance of these models by first finetuning them for a single epoch on MS-MARCO triples \citep{bajaj2018msmarco}, using the standard InfoNCE loss. We then evaluate three retrieval tasks- Natural Questions \citep{kwiatkowski-etal-2019-natural-questions} for general-purpose IR, Code Search Net \citep{husain2019codesearchnet} for code retrieval, and MLDR \citep{chen2024bgem3} for long-context retrieval.
\end{itemize}

For all tasks, we compare the base Granite Encoder model with popular open-source base-sized encoders such as BERT \citep{devlin2019bert}, RoBERTa \citep{liu2019roberta}, and ModernBERT \citep{warner2024modernbert}. The small Granite Encoder is compared against the 12-layer MiniLM model \citep{wang2020minilm}, which matches its architecture exactly, and is a popular starting point for training smaller embedding models. We evaluate all the models with the same pipeline to ensure fair comparison.

As shown in Table~\ref{tab:encoder-performance}, the Granite Encoder models show strong performance on both NLU and retrieval tasks, while being trained on high-quality data suitable for enterprise use.

\section{Detailed Retriever Performance Evaluation}
\label{app:evals}
\subsection{English Text Embedding Performance}
\label{app:mteb-v2}

\begin{table}[!ht]
    \centering
    \begin{adjustbox}{width=1\textwidth}
    {\renewcommand{\arraystretch}{1.3}
    \begin{tabular}{l|c|ccccccc}
    \toprule
        Task                               & Avg. & Class. & Clustering & PairClass. & Reranking & Retrieval & STS & Summ. \\ 
        \# of datasets $\rightarrow$  & 41 & 8 & 8 & 3 & 2 & 10 & 9 & 1 \\ 
        \hline
        e5-base-v2                          & 62.7 & 75.5 & 45.2 & 85.6 & 45.1 & 49.7 & 80.6 & 34.3 \\ 
        bge-base-en-v1.5                    & 65.1 & 77.7 & 47.4 & 86.6 & 46.7 & 54.8 & 82.1 & 30.2 \\ 
        snowflake-arctic-embed-m-v2.0       & 62.6 & 71.4 & 43.6 & 82.3 & 46.9 & 58.4 & 76.6 & 30.3 \\ 
        gte-base-en-v1.5 & 65.8 & 81.0 & 47.9 & 85.4 & 46.6 & 55.5 & 81.7 & 29.0 \\ 
        gte-modernbert-base & 66.3 & 80.9 & 47.8 & 86.6 & 47.6 & 57.0 & 81.5 & 31.8 \\ 
        nomic-ai/modernbert-embed-base & 62.1 & 75.1 & 46.3 & 82.4 & 46.3 & 48.7 & 79.3 & 34.1 \\ 
        e5-small-v2                         & 61.3 & 74.6 & 42.1 & 84.8 & 44.1 & 48.5 & 80.1 & 33.2 \\ 
        bge-small-en-v1.5                   & 64.3 & 76.6 & 47.0 & 85.0 & 46.9 & 53.9 & 81.3 & 28.9 \\ 
        \midrule
        granite-embedding-125m-english      & 62.1 & 68.3 & 47.3 & 79.6 & 49.4 & 55.7 & 77.6 & 29.3 \\ 
        granite-embedding-30m-english       & 60.2 & 67.7 & 46.0 & 78.9 & 47.5 & 51.8 & 75.9 & 24.2 \\ 
        \textbf{granite-embedding-english-r2}  & 62.8 & 70.7 & 47.2 & 79.3 & 49.1 & 56.4 & 78.1 & 29.3 \\ 
        \textbf{granite-embedding-small-english-r2}  & 61.1 & 68.2 & 46.6 & 78.6 & 47.6 & 53.9 & 76.9 & 26.7 \\ 
    \bottomrule
    \end{tabular}}
    \end{adjustbox}
    \caption{MTEB-v2 Benchmark \cite{enevoldsen2025mmtebmassivemultilingualtext} Performance. We show the average score per task using the main metric described in the paper. Other than the Granite Embedding R2 models all results have been obtained from the public MTEB-v2 English leaderboard  as of August 8, 2025.\textit{Class., Summ.} is short for \textit{Classification, Summarization}, respectively.}
\label{tab:mteb-v2}
\end{table}

For general-purpose embedding performance beyond retrieval, we measure the performance of the Granite Embedding Models on the English MTEB-v2 benchmark, which consists of 7 tasks spanning 41 datasets, used to evaluate the quality of text embeddings on classification, clustering, pair classification, reranking, retrieval, semantic similarity and summarization.

We provide the average score per task in Table \ref{tab:mteb-v2}, using the main metric described in the paper: Accuracy for classification tasks, V-Measure for clustering tasks, Average Precision for pair classification tasks, MAP for re-ranking tasks, nDCG@10 for retrieval tasks and Spearman Correlation (based on cosine similarity) for STS and summarization tasks.

While Granite Embedding models are purpose built for Retrieval tasks, they still maintain a strong performance on the other tasks of the MTEB benchmark, notably without adding any training data from non-retrieval tasks. This indicates the high quality of the embeddings produced by these models.

\subsection{English Retrieval Performance}

\begin{table}[!ht]
    \centering
    \begin{adjustbox}{width=1.0\textwidth}
    {\renewcommand{\arraystretch}{1.3}
    \begin{tabular}{l|c|ccccccccccccccc}
    \toprule
        Model                               & Avg. & Argu- & CQA & Climate & DB- & Fever & FiQA & Hotpot & MS & NF & NQ & Quora & Sci & Sci & Trec & Touche \\ 
         &  & Ana &  & Fever & Pedia &  &  & QA & Marco & Corpus &  &  & Docs & Fact & Covid & \\  
        \hline

        e5-base-v2                          & 50.3 & 44.6 & 38.5 & 26.6 & 42.2 & 85.0 & 39.9 & 69.2 & 41.8 & 35.4 & 58.2 & 86.6 & 18.7 & 71.9 & 69.6 & 26.4 \\ 
        bge-base-en-v1.5                    & 53.2 & 63.8 & 41.6 & 31.2 & 40.8 & 86.3 & 40.6 & 72.6 & 41.4 & 37.4 & 54.1 & 88.9 & 21.7 & 74.3 & 78.0 & 25.7 \\ 
        snowflake-arctic-embed-m-v2.0       & 55.5 & 57.9 & 47.2 & 38.1 & 43.9 & 91.7 & 44.2 & 72.4 & 44.0 & 35.9 & 64.6 & 89.1 & 20.3 & 72.3 & 80.3 & 30.4 \\ 
        gte-base-en-v1.5 & 54.1 & 63.5 & 39.5 & 40.4 & 39.9 & 94.8 & 48.7 & 67.8 & 42.6 & 35.9 & 53.0 & 88.4 & 21.9 & 76.8 & 73.1 & 25.2 \\ 
        gte-modernbert-base & 55.2 & 74.6 & 42.6 & 45.9 & 41.4 & 94.0 & 49.5 & 70.4 & 39.9 & 34.3 & 56.1 & 88.6 & 20.4 & 76.4 & 75.7 & 18.0 \\ 
        nomic-ai/modernbert-embed-base & 52.9 & 49.0 & 42.1 & 35.7 & 41.5 & 87.4 & 40.6 & 67.1 & 41.5 & 33.4 & 62.2 & 88.8 & 18.6 & 69.6 & 84.1 & 31.9 \\ 
        e5-small-v2                         & 49 & 41.8 & 37.1 & 22.9 & 41.3 & 81.6 & 37.4 & 66.6 & 41.5 & 32.4 & 59.1 & 85.7 & 17.7 & 68.9 & 74.4 & 27.1 \\ 
        bge-small-en-v1.5                   & 51.7 & 60.3 & 38.3 & 31.8 & 40.0 & 86.6 & 40.3 & 69.9 & 40.8 & 34.3 & 50.2 & 88.8 & 20.5 & 71.3 & 75.5 & 26.0 \\ \midrule
        granite-embedding-125m-english      & 52.3 & 58.4 & 48.2 & 33.2 & 39.4 & 88.2 & 44.9 & 67.8 & 32.5 & 37.3 & 58.1 & 87.8 & 24.2 & 74.7 & 69.3 & 20.2 \\ 
        granite-embedding-30m-english       & 49.1 & 56.4 & 44.3 & 30.3 & 36.0 & 85.5 & 36.9 & 62.9 & 30.7 & 33.7 & 51.6 & 86.7 & 22.5 & 71.3 & 63.1 & 24.0 \\ 
        \textbf{granite-embedding-english-r2}  & 53.1 & 59.2 & 50.0 & 35.8 & 39.6 & 88.1 & 46.3 & 67.4 & 32.1 & 37.6 & 58.2 & 87.8 & 25.0 & 75.8 & 70.6 & 22.9 \\ 
        \textbf{granite-embedding-small-english-r2}  & 50.9 & 54.4 & 47.9 & 31.6 & 37.9 & 86.5 & 40.9 & 65.7 & 30.1 & 37.1 & 55.4 & 87.4 & 24.0 & 75.5 & 64.6 & 24.3 \\ 
    \bottomrule
    \end{tabular}}
    \end{adjustbox}
    \caption{BEIR retrieval benchmark performance. All scores are NDCG@10.}
    \label{tab:beir}
\end{table}

\begin{table}[!ht]
    \centering
    \begin{adjustbox}{width=1.0\textwidth}
    {\renewcommand{\arraystretch}{1.3}
    \begin{tabular}{l|c|cccccccccc}
    \toprule
        Model                               & Avg. & Argu- & CQA  & CQA  & Climate & Fever & FiQA & Hotpot & Sci- & Touche & Trec \\ 
        &  & Ana &  Gaming &  Unix & Fever &  &  & QA & Docs &  & Covid \\ \hline
        e5-base-v2                          & 49.7 & 44.6 & 55.7 & 36.9 & 27.3 & 85.0 & 39.9 & 69.1 & 18.7 & 69.6 & 49.9 \\ 
        bge-base-en-v1.5                    & 54.8 & 63.8 & 59.6 & 42.2 & 32.1 & 86.0 & 40.6 & 73.7 & 21.7 & 78.0 & 49.7 \\ 
        snowflake-arctic-embed-m-v2.0       & 58.4 & 57.9 & 65.3 & 47.8 & 38.7 & 92.3 & 44.2 & 71.9 & 20.3 & 80.3 & 65.4 \\ 
        gte-base-en-v1.5 & 55.5 & 63.7 & 57.3 & 37.2 & 40.9 & 95.0 & 48.8 & 69.7 & 21.9 & 47.2 & 73.0 \\ 
        gte-modernbert-base & 57.0 & 74.4 & 60.5 & 43.3 & 46.3 & 94.7 & 52.5 & 71.3 & 19.5 & 44.3 & 63.3 \\ 
        nomic-ai/modernbert-embed-base & 48.7 & 50.5 & 59.5 & 40.7 & 27.9 & 70.0 & 39.1 & 55.9 & 20.0 & 55.6 & 67.6 \\ 
        e5-small-v2                         & 48.5 & 41.8 & 55.7 & 36.0 & 23.2 & 81.7 & 37.4 & 66.9 & 17.7 & 74.4 & 49.8 \\ 
        bge-small-en-v1.5                   & 53.9 & 60.3 & 56.4 & 38.8 & 32.2 & 86.2 & 40.3 & 70.7 & 20.5 & 75.5 & 57.6 \\ \midrule
        granite-embedding-125m-english      & 55.7 & 58.4 & 63.6 & 51.3 & 33.1 & 90.0 & 44.9 & 68.1 & 24.2 & 53.6 & 69.3 \\ 
        granite-embedding-30m-english       & 51.8 & 56.4 & 59.3 & 46.1 & 30.6 & 86.5 & 36.9 & 63.1 & 22.5 & 53.8 & 63.1 \\ 
        \textbf{granite-embedding-english-r2}  & 56.4 & 59.2 & 65.0 & 52.9 & 36.2 & 88.8 & 46.2 & 67.1 & 24.9 & 52.8 & 70.3 \\ 
        \textbf{granite-embedding-small-english-r2}  & 53.9 & 54.5 & 62.5 & 51.2 & 31.5 & 87.5 & 40.7 & 66.2 & 24.0 & 56.2 & 64.7 \\ 
    \bottomrule
    \end{tabular}}
    \end{adjustbox}
    \caption{MTEB v2 retrieval Performance. All scores are NDCG@10.}
    \label{tab:mteb-v2-retrieval}
\end{table}

We measure the performance of our models on two popular retrieval benchmarks:
\begin{itemize}
    \item BEIR \citep{thakur2021beir}: This Information Retrieval benchmark consist of 15 datasets spanning multiple domains, and is used to test the model's ability to find the relevant document for a given query. The tasks of this benchmark also comprise the MTEB v1 \citep{muennighoff2022mteb} retrieval tasks.
    \item MTEB v2 retrieval \citep{enevoldsen2025mmtebmassivemultilingualtext}: this benchmark is an update to the MTEB v1 retrieval benchmark, wherein they exclude tasks such as MS-Marco and NQ which are often used in the finetuning of embedding models.
\end{itemize}

The experiment results for the two tasks are reported in Table \ref{tab:beir} and Table \ref{tab:mteb-v2-retrieval}. We report the nDCG@10 scores on each dataset, showing strong performance relative to other open-source models of similar sizes. We also note the average performance on all tasks except MS-MARCO retrieval \citep{bajaj2018msmarco}, for a fair zero-shot comparison, as other embedding models train on MS-MARCO, but our models do not, due to unfavorable license. Despite being trained on less data, and only permissibly licensed public datasets, our models show strong performance.

Note, other than the Granite Embedding R2 models, all model performances have been reported from the MTEB leaderboards as of August 8th 2025.

\subsection{Code Retrieval Performance}

\begin{table}[!ht]
    \centering
    \begin{adjustbox}{width=1\textwidth}
    {\renewcommand{\arraystretch}{1.3}
    \begin{tabular}{l|c|cccccccccc}
    \toprule
    Model                               & Avg. & Apps & COIR & CFB & CFB & CSN & CT & CT & CosQA & Stack & Syn. \\
    & &  & CSN & MT & ST & CCR & Contest & DL &  & OverFlow QA & Text2sql \\ \hline
    e5-base-v2                          & 50.3 & 11.5 & 62.7 & 41.6 & 74.5 & 56.9 & 62.5 & 21.9 & 32.6 & 87.8 & 51.9 \\ 
        bge-base-en-v1.5                    & 46.6 & 6.5 & 72.0 & 33.6 & 70.0 & 50.9 & 45.6 & 23.5 & 33.7 & 80.2 & 50.0 \\ 
        snowflake-arctic-embed-m-v2.0       & 52.2 & 10.8 & 51.8 & 45.3 & 77.6 & 52.8 & 75.0 & 28.2 & 36.8 & 90.7 & 52.9 \\ 
        gte-base-en-v1.5 & 42.4 & 5.3 & 31.1 & 47.8 & 65.3 & 35.0 & 57.4 & 27.3 & 32.4 & 81.2 & 40.7 \\ 
        gte-modernbert-base & 71.5 & 56.4 & 83.1 & 85.8 & 85.5 & 93.4 & 73.0 & 36.1 & 42.2 & 90.9 & 64.7 \\ 
        nomic-ai/modernbert-embed-base & 48.8 & 5.2 & 72.8 & 64.9 & 41.1 & 56.5 & 56.3 & 29.7 & 34.4 & 78.5 & 48.1 \\ 
        e5-small-v2                         & 47.1 & 4.4 & 55.5 & 39.8 & 71.2 & 51.3 & 53.3 & 30.8 & 29.7 & 83.5 & 52.1 \\ 
        bge-small-en-v1.5                   & 45.8 & 5.6 & 72.6 & 35.1 & 67.8 & 47.9 & 48.2 & 25.7 & 32.0 & 78.0 & 45.1 \\ \midrule
    granite-embedding-125m-english      & 50.3 & 11.8 & 55.1 & 42.1 & 75.3 & 47.6 & 66.7 & 29.6 & 36.6 & 89.8 & 48.7 \\ 
    granite-embedding-30m-english       & 47.0 & 6.2 & 58.7 & 37.8 & 69.4 & 49.1 & 57.5 & 26.9 & 35.4 & 83.9 & 44.4 \\ 
    \textbf{granite-embedding-english-r2}  & 54.8 & 14.0 & 65.7 & 52.5 & 77.2 & 47.7 & 77.1 & 35.0 & 37.0 & 91.8 & 49.6 \\ 
    \textbf{granite-embedding-small-english-r2}  & 53.4 & 13.5 & 60.4 & 52.2 & 76.9 & 48.4 & 77.6 & 33.6 & 35.4 & 90.0 & 46.3 \\ 
    \bottomrule
    \end{tabular}}
    \end{adjustbox}
    \caption{Code Retrieval Performance on COIR. All scores are NDCG@10. \textit{CSN, CFB,
CT} is short for \textit{CodeSearchNet, CodeFeedBack, CodeTrans}, respectively.}
\label{tab:coir}
\end{table}

We evaluate the model's code retrieval capability on the COIR benchmark \citep{li2024coircomprehensivebenchmarkcode}, consisting of 10 datasets across 7 domains in Table \ref{tab:coir}. Granite Embedding models show strong performance compared models of the same size, despite the fact that unlike other models, we do not include any COIR training data in the training of our models, leading to purely zero-shot evaluation for our models.

\subsection{Long-Context Retrieval Performance}

\begin{table}[ht!]
\begin{adjustbox}{width=1\textwidth}
{\renewcommand{\arraystretch}{1.3}
\begin{tabular}{l|c|c|ccccccc}
\toprule
Model & MSL & MLDR & \multicolumn{7}{c}{LongEmbed} \\
 &  & & NQA & Needle & PassKey & QMSum & SFD & 2WmQA & Avg. \\
\hline
e5-base-v2 & 512 & 32.5 & 25.3 & 28.5 & 38.0 & 23.9 & 74.7 & 56.0 & 41.1 \\
bge-base-en-v1.5 & 512 & 33.5 & 25.6 & 25.3 & 18.0 & 22.4 & 60.3 & 51.7 & 33.9 \\
snowflake-arctic-embed-m-v2.0 & 8192 &  32.4 & 36.2 & 39.5 & 58.0 & 29.8 & 93.4 & 75.4 & 55.4 \\
gte-base-en-v1.5 & 8192 & 42.7 & 47.3 & 31.8 & 58.0 & 41.7 & 92.1 & 85.6 & 59.4 \\
gte-modernbert-base & 8192 &  46.2 & 30.2 & 24.5 & 66.8 & 40.7 & 93.1 & 87.0 & 57.0 \\
nomic-ai/modernbert-embed-base & 8192 &  31.3 & 41.1 & 35.8 & 62.5 & 36.7 & 94.5 & 67.5 & 56.3 \\
e5-small-v2 & 512 &  29.9 & 25.1 & 29.5 & 38.0 & 24.1 & 69.7 & 58.0 & 40.7 \\
bge-small-en-v1.5 & 512 &  31.4 & 22.2 & 26.8 & 21.0 & 21.3 & 57.8 & 43.5 & 32.1 \\
\midrule
granite-embedding-125m-english & 512 &  35.0 & 25.4 & 32.0 & 38.8 & 25.2 & 71.4 & 60.6 & 42.2 \\
granite-embedding-30m-english & 512 &  32.6 & 21.7 & 29.8 & 38.5 & 23.8 & 60.2 & 56.7 & 38.4 \\
\textbf{granite-embedding-english-r2} & 8192 &  41.6 & 49.7 & 49.8 & 83.3 & 42.8 & 94.2 & 86.9 & 67.8 \\
\textbf{granite-embedding-small-english-r2} & 8192 & 40.1 & 41.4 & 55.0 & 68.5 & 36.5 & 89.9 & 79.9 & 61.9 \\
\bottomrule
\end{tabular}}
\end{adjustbox}
\caption{Long Context Retrieval Performance on MLDR and LongEmbed. MSL depicts the maximum sequence length of the embedding model. All scores are NDCG@10 except the Needle and Passkey subsets, which report Accuracy@1. \textit{NQA, SFD,
2WmQA} is short for \textit{NarrativeQA, SummScreenFD, 2WikiMultihopQA}, respectively.}
\label{tab:longctx}
\end{table} 

For evaluating the performance of our models on long-context documents, we measure performance on the following benchmarks:
\begin{itemize}
    \item LongEmbed: The LongEmbed benchmark contains two synthetic and four real-world tasks, designed to benchmark long context retrieval. 
    \item MLDR (English): Multilingual Long-Document Retrieval dataset is built on Wikipeida, Wudao and mC4, covering 13 languages, with questions generated using GPT-3.5. We limit our evaluations to the English subset of this dataset
\end{itemize}

As shown in Table \ref{tab:longctx}, the Granite Embedding R2 models show very strong performance on the long context benchmarks, with state-of-the-art performance on LongEmbed compared to other models trained with long context.

\subsection{Table Retrieval Performance}

\begin{table}[!ht]
    \centering
    \begin{adjustbox}{width=0.9\textwidth}
    {\renewcommand{\arraystretch}{1.3}
    \begin{tabular}{l|c|c|c|c|c|c}
    \toprule
        Model                               & Avg. & OpenWikiTables & NQTables & OTT-QA & MultiHierTT & AIT-QA \\ 
        & & R@5 & R@5 & R@5 & M@5 & M@5 \\
        \hline
        e5-base-v2                          & 74.1 & 94.9 & 70.9 & 92.6 & 60.7 & 51.3 \\ 
        bge-base-en-v1.5                    & 74.0 & 95.1 & 68.5 & 93.6 & 55.7 & 57.1 \\ 
        snowflake-arctic-embed-m-v2.0       & 80.8 & 97.6 & 61.4 & 94.3 & 71.3 & 79.2 \\ 
        gte-base-en-v1.5                    & 80.5 & 95.9 & 72.4 & 93.0 & 69.1 & 72.2 \\ 
        gte-modernbert-base                 & 76.7 & 92.8 & 70.3 & 93.9 & 67.2 & 59.2 \\ 
        nomic-ai/modernbert-embed-base      & 66.7 & 89.6 & 58.7 & 86.9 & 44.6 & 53.8 \\ 
        e5-small-v2                         & 72.3 & 94.8 & 61.9 & 91.7 & 59.5 & 53.6 \\ 
        bge-small-en-v1.5                   & 69.9 & 94.2 & 65.4 & 92.4 & 53.3 & 44.3 \\ 
        \midrule
        granite-125m-english                & 73.3 & 96.4 & 74.8 & 94.7 & 52.9 & 47.6 \\ 
        granite-embedding-30m-english       & 69.2 & 96.5 & 71.1 & 93.0 & 46.5 & 38.8 \\ 
        \textbf{granite-embedding-english-r2}  & 78.5 & 97.2 & 75.2 & 94.9 & 58.3 & 67.0 \\ 
        \textbf{granite-embedding-small-english-r2}   & 75.5 & 97.3 & 72.0 & 94.1 & 52.4 & 61.8 \\ 
    \bottomrule
    \end{tabular}}
    \end{adjustbox}
    \caption{Retrieval performance comparison across 5 Table-IR datasets. We report Recall@5 (R@5) for \textit{OpenWikiTables}, \textit{NQTables}, \textit{OTT-QA}, and Match@5 (M@5) for \textit{MultiHierTT} and \textit{AIT-QA}. All the tables are in markdown format.}
\label{tab:tableir_performance}
\end{table}

 To evaluate the performance of our Granite Embedding R2 models on tabular retrieval tasks, we assess them using the following datasets:

\begin{itemize}
    \item OpenWikitables: An open-domain QA dataset carefully designed for Table-IR task in open domain setting, derived from WikiSQL and WikiTableQuestions.

    \item NQTables: A large-scale table retrieval subset extracted from the Natural Questions dataset, focusing on open-domain retrieval and question answering over Wikipedia tables.

    \item OTT-QA: Open domain table-and-text question answering dataset requiring retrieval and fusion of both tabular data and textual passages from large pools. We use the corpus and queries included in the TARGET Benchmark \citep{ji2024target} for evaluation.

    \item MultiHierTT: A benchmark for numerical reasoning over multi-hierarchical tabular and textual data extracted from financial reports. The documents contain multiple hierarchical tables and lengthy narrative text, requiring complex multi-step reasoning. As many queries are context-dependent, we decontextualize them prior to evaluation in the retrieval task.

    \item AIT-QA: A domain-specific table QA dataset in the airline industry, comprising 515 human-annotated questions over 116 complex, hierarchical tables sourced from SEC filings. We use an extended version of AIT-QA that incorporates both the tables and their source 10-K forms from the SEC, enabling a hybrid retrieval setting that utilizes tabular and textual data. The final version of the evaluation set consists of 515 questions with ground truth answers and pages with a total search corpus of 1939 pages with 1682 tables.
\end{itemize}

As shown in Table~\ref{tab:tableir_performance}, the Granite Embedding R2 models demonstrate consistently strong retrieval performance across all Table-IR datasets.

\section{RoPE Theta scaling Ablations}
\label{app:rope-theta}

To investigate the impact of global RoPE theta, we conducted an ablation study on the granite-embedding-small-english-r2 model using values of 20k, 40k, 80k, and 160k. In the ModernBERT architecture, where global and local attention layers alternate, we adjusted RoPE theta only for the global attention layers, keeping local attention fixed at the default 10k. All experiments were conducted during the final distillation stage with the same setup as discussed in Section \ref{subsec: general training recipe}.

As shown in the Table~\ref{tab:rope}, benchmarks such as MTEB-v1 and CoIR were insensitive to changes in global RoPE theta, whereas MLDR achieved its highest performance at 80k. Increasing the value to 160k did not yield consistent improvements across benchmarks. For reference, inference results with a global RoPE theta of 160k are also reported for comparison with the best performing 80k configuration.

\begin{table}[!t]
    \centering
    \setlength{\tabcolsep}{0.5em}
    {\renewcommand{\arraystretch}{1.3}
    \begin{tabular}{l|c|c|c}
    \toprule
        Global RoPE Theta                   &  MTEB-v1 &  CoIR &  MLDR \\ 
        & Retrieval (15)  &  (10)  & (En) \\
        \hline
        20k       & 50.9 & 53.4 & 38.6 \\ 
        40k       & 50.9 & 53.4 & 39.6 \\ 
        \textbf{80k}       & \textbf{50.9} & \textbf{53.4} & \textbf{40.1} \\ 
        160k      & 50.9 & 53.6 & 39.3 \\ 
        \midrule
        80k with 160k inference  & 50.9 & 53.8 & 39.4 \\ 
    \bottomrule
    \end{tabular}}
    \caption{Global RoPE Theta Scaling Performance. The top four rows display the global RoPE theta values tested during training. The final row presents inference scaling result for the best performing model (80k during training).}
\label{tab:rope}
\end{table}

\section{Retriever Training Hyperparameters}
\label{app:hyperparameters}

\begin{table}[H]
    \centering
    \setlength{\tabcolsep}{0.5em}
    \begin{adjustbox}{width=1.0\textwidth}
    {\renewcommand{\arraystretch}{1.3}
    \begin{tabular}{l|c|cccc}
    \toprule
        Model & Stage & LR & Batch Size & Steps & Seq. Len\\
        \hline
        \multirow{5}{*}{granite-embedding-english-r2} & RetroMAE PT &  $2e^{-5}$ & $8$ & $55864$ & $8192$ \\
        & Tabular PT & $2e^{-6}$ & $1024$ & $8848$ & $2048$  \\
        & Contrastive FT & $2e^{-4}$ & $9600$ & $58000$ & $512$ \\
        & Contrastive FT (w/ HN) & $4e^{-6}$ & $450$ & $600$ & $1024$ \\
        & Contrastive KD & $1e^{-5}$ & $256$ & $7000$ & $512$ \\
        & Domain Adaptation & $5e^{-6}$ &$256$ &$100$& $512$ \\
        \midrule
        \multirow{3}{*}{granite-embedding-small-english-r2} & RetroMAE PT &  $2e^{-5}$ & $16$ & $55864$ & $8192$\\
        & Contrastive FT & $7e^{-4}$ & $12000$ & $70000$ & $512$ \\
        & Contrastive KD & $2e^{-5}$ & $32$ & $12000$ & $1024$ \\
    \bottomrule
    \end{tabular}}
    \end{adjustbox}
    \caption{Retriever Training Hyperparameters. Batch size refers to the global batch size. PT, FT, KD refers to Pre-training, finetuning and knowledge distillation respectively.}
\label{tab:hps}
\end{table}

Table~\ref{tab:hps} provides the detailed hyperparameters for each stage of retriever training for the Granite Embedding models.

\end{document}